\documentclass[letterpaper, 10 pt, conference]{ieeeconf}  % Comment this line out if you need a4paper

\IEEEoverridecommandlockouts                              % This command is only needed if 
\usepackage{graphicx, tipa}
\usepackage{amsmath,amssymb,amsfonts}
\usepackage[caption=false]{subfig}
\usepackage[ruled,linesnumbered, noend]{algorithm2e}
\usepackage{dirtytalk}
\usepackage{hyperref}
\usepackage{gensymb}
\usepackage{tabularx}
\usepackage{multirow}
\usepackage[short]{optidef}
\usepackage{booktabs}
\usepackage{siunitx}
\usepackage{titlesec}
\usepackage{xcolor}

\title{\LARGE \bf
ASIP-Planner: Adaptive Planning for UAV Surface Inspection in Partially Known Indoor Environments
}

\author{Hanyu Jin$^{1}$, Zhefan Xu$^{1}$, Haoyu Shen$^{1}$, Xinming Han$^{2}$, Kanlong Ye$^{1}$, and Kenji Shimada$^{1}$%
\thanks{$^{1}$Hanyu Jin, Zhefan Xu, Haoyu Shen, Kanlong Ye, and Kenji Shimada are with the Department of Mechanical Engineering, Carnegie Mellon University, Pittsburgh, PA 15213, USA. {\tt\footnotesize hanyujin@andrew.cmu.edu}}%
\thanks{$^{2}$Xinming Han is with the Department of Mechanical and Aerospace Engineering, University of California, Davis, Davis, CA 95616, USA.}%
}

\begin{document}

\maketitle
\thispagestyle{empty}
\pagestyle{empty}

%%%%%%%%%%%%%%%%%%%%%%%%%%%%%%%%%%%%%%%%%%%%%%%%%%%%%%%%%%%%%%%%%%%%%%%%%%%%%%%%
\begin{abstract}
Indoor infrastructure inspection, such as tunnels and industrial facilities, requires systematic surface coverage to ensure that all inspection targets are properly observed.
Unmanned Aerial Vehicles (UAVs) offer an alternative to manual inspection by conducting map-guided surface inspection using prior structural models. 
However, in practice, indoor inspection often relies on floorplan-derived reference maps that may not reflect unforeseen obstacles, such as temporary structures or equipment, leading to occluded viewpoints and degraded inspection quality.
Existing coverage planning methods typically assume a fully known inspection environment and perform deterministic global viewpoint optimization based on accurate prior maps, making them vulnerable to environmental discrepancies during execution.
This work presents an adaptive UAV inspection framework for partially known structured indoor environments. 
The proposed method integrates a segment-based global coverage planner with an inspection-oriented local view-angle adaptation module. 
The global planner organizes planar inspection targets into surface-aligned clusters to generate compact viewpoint sequences with improved orientation consistency. 
During execution, the local planner generates collision-free trajectories and adjusts the viewing direction online to mitigate occlusion-induced coverage loss while preserving the planned trajectory structure. 
The simulation results across randomized scene configurations demonstrate that the proposed global planner achieves near-complete coverage while reducing trajectory length and mean absolute yaw change compared to representative baselines.
Real-world flight experiments further validate that the framework can be executed in indoor environments and produce usable inspection data for downstream analysis.
These results indicate that the proposed framework improves inspection efficiency and adaptability in partially known structured indoor environments.
\end{abstract}

%%%%%%%%%%%%%%%%%%%%%%%%%%%%%%%%%%%%%%%%%%%%%%%%%%%%%%%%%%%%%%%%%%%%%%%%%%%%%%%%
\section{INTRODUCTION}
Indoor inspection has been a critical problem for infrastructure maintenance. Manual inspection in such environments is labor-intensive and potentially hazardous, motivating the use of Unmanned Aerial Vehicles (UAVs) for autonomous inspection. To achieve high-quality inspection and reconstruction, UAVs must generate trajectories that provide consistent and comprehensive coverage of interior surfaces while satisfying visibility, safety, and motion constraints. In many indoor scenarios derived from architectural floor plans, inspection targets exhibit strong geometric regularity, often consisting of vertically extruded planar surfaces. Such structures motivate systematic coverage planning to ensure consistent observation of interior surfaces.

Existing UAV inspection methods typically rely on reference maps to define inspection targets and generate global coverage trajectories \cite{Feng2024FCPlanner}\cite{HCPP}\cite{Wang2025Uav3DSurfaceInspection}. Under the assumption of accurate prior knowledge, these approaches plan viewpoint sequences to minimize travel distance while ensuring surface visibility. 
However, in partially known indoor environments, the reference map may be incomplete or inconsistent with the real scene due to unmodeled obstacles such as temporary structures or equipment. 
As a result, pre-planned viewpoints may become occluded or unreachable during execution, leading to coverage degradation and incomplete observation of inspection targets.

\begin{figure}[t]
  \centering
  \includegraphics[width=\linewidth]{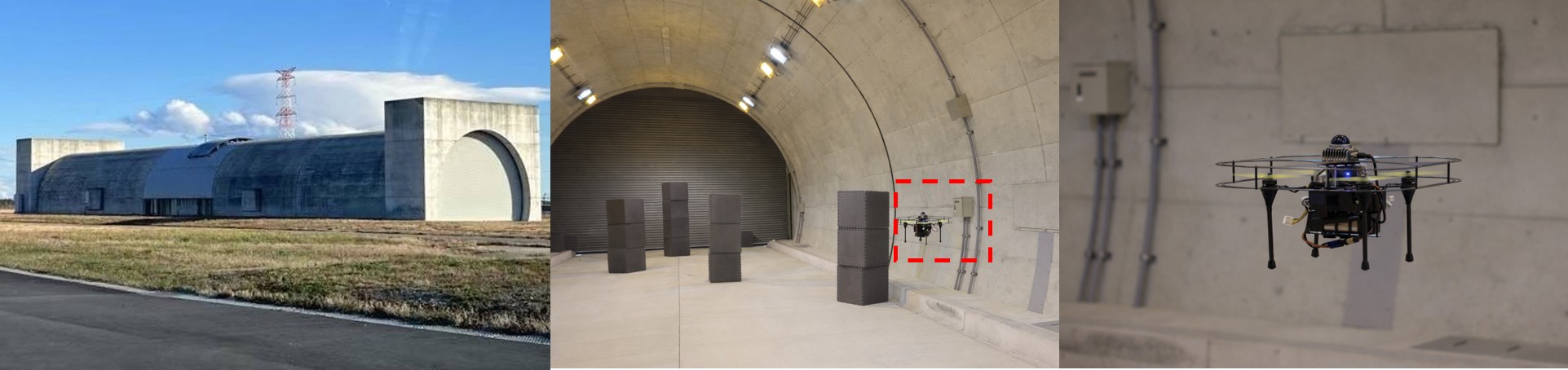}
  \caption{UAV-based surface inspection in a tunnel infrastructure. 
  Left: Exterior view of the tunnel in Fukushima Robot Test Field \cite{TestFieldWebsite}. 
    Middle: Interior view of the tunnel environment with obstacles used to simulate partial environmental knowledge during inspection. 
    Right: Close-up view of our UAV inspecting the tunnel wall surface.
    % Left: Our UAV is inspecting interior tunnel surfaces in the Fukushima Robot Test Field. 
    % Right: Exterior view of the inspected tunnel \cite{TestFieldWebsite}.
    }
  \label{fig:cover}
\end{figure}

To address these challenges, we propose an Adaptive planning framework for Surface Inspection in Partially known indoor environments (ASIP-Planner). The framework consists of two tightly coupled components:  
(1) a segment-based global planner organizes the structured inspection targets into coherent viewpoint clusters, enabling compact sequencing and consistent inspection execution, and (2) a local planner operates during execution to adapt the UAV trajectory and viewing direction in response to unforeseen obstacles, mitigating coverage loss due to occlusions or environmental discrepancies. 
By combining structured global coverage planning with online local adaptation, the proposed system maintains inspection efficiency while improving coverage reliability under environmental uncertainty. Figure \ref{fig:cover} shows the deployment of our framework in tunnel interior wall inspection. 

The main contributions of this work include:
\begin{itemize}
    \item \textbf{Surface Inspection Planning Framework: }We propose a UAV inspection planning framework that combines segment-based global coverage planning with adaptive local view-angle adjustment for inspection in partially unknown indoor environments.
    \item \textbf{Segment-based Global Coverage Planning: }We develop a segment-based viewpoint generation and sequencing strategy that organizes inspection targets into groups defined by surface normals, resulting in compact viewpoint clusters, reduced interleaving between surfaces, and improved execution consistency.
    \item \textbf{Inspection-oriented Local Adaptive Planning: }We design a local planning framework that integrates adaptive view-angle selection with trajectory execution. The view-angle planner dynamically adjusts the viewing direction to compensate for occlusions and improve coverage in the presence of environmental uncertainty.
    \item \textbf{Real-world Deployment and Application Validation: }We validate the proposed framework through autonomous UAV flight experiments in multiple indoor environments, including a tunnel inspection scenario. We further demonstrate the applicability of inspection data to downstream tasks by conducting post-processing such as 3D reconstruction and defect detection.
    
\end{itemize}

% outdoor inspection \cite{road_inspection}
% main contribution: 
% \begin{itemize}
% \item autonomous inspection framework in partially unknown environment 
% \item global viewpoint planner
% \item local view angle adaptation
% \end{itemize}
\section{RELATED WORK}
Autonomous UAV inspection methods can be broadly categorized based on the availability of prior environmental information. In fully unknown environments, inspection is typically treated as an exploration problem, in which the UAV incrementally reconstructs the scene while navigating and expanding the observable space \cite{Zhang2025FALCON}\cite{Dhami2023PredNBV}\cite{Burusa2024GradientNBV}\cite{Xue2024NeuralVisibilityField}. These approaches focus on reducing environmental uncertainty and discovering unseen regions, but they do not explicitly optimize inspection-oriented objectives such as surface completeness or task-specific viewpoint quality. 

In contrast, inspection in fully known environments assumes access to an accurate geometric model (e.g., CAD, mesh, or BIM) and formulates the problem as deterministic coverage path planning (CPP). The pipeline typically consists of two stages: (i) generating inspection viewpoints that satisfy visibility, distance, and sensing constraints, and (ii) solving a global path optimization problem to connect viewpoints efficiently under UAV motion and safety constraints. Recent works have improved viewpoint sampling efficiency and coverage quality through multi-objective risk-aware optimization \cite{Petit2024MoarPlanner}, structural inspection optimization \cite{Zhao2024OptimizedStructuralInspection}, skeleton-guided scene decomposition \cite{Feng2024FCPlanner}, and surface-normal-based viewpoint evaluation metrics \cite{Wang2025Uav3DSurfaceInspection}. Learning-based inspection planning has also been explored to infer inspection policies from demonstration \cite{Kannan2023UPPLIED}, while BIM-assisted frameworks leverage detailed structural priors to improve coverage efficiency in structured indoor environments \cite{Chen2024ImprovedCoverageBIM}. Multi-robot extensions further address task allocation and coordinated coverage assuming complete environmental knowledge \cite{Wang2025MACPlanner}. Although these methods achieve efficient and high-quality coverage under strong prior assumptions, they rely on accurate and complete models, and thus are not directly applicable when the prior map is incomplete or partially unreliable. Unlike previous works, our problem formulation focuses on interior planar surfaces derived from structured floorplans. By leveraging the geometric regularity of such environments, we organize viewpoints into coherent segments and promote consistent viewing directions during inspection.

Between fully known inspection and pure exploration lies the practical setting of partially known environments, where prior maps are available but incomplete or inconsistent with the real scene. In such cases, the inspection objective must be achieved despite structural uncertainty and unforeseen obstacles. Xu et al. \cite{Xu2023VisionBasedTunnelInspection} addressed inspection in unknown tunnel environments by integrating perception and reactive planning to handle dynamic obstacles; however, their method does not leverage a global coverage plan derived from prior structural information. Gao et al. \cite{Gao2023ExploreThenExploit} proposed an explore-then-exploit framework that first reconstructs the environment before performing coverage planning, which introduces additional exploration overhead and decouples mapping from inspection optimization. Dhami et al. \cite{dhami2024gatsbionlinegtspbasedalgorithm} formulated online inspection planning as a GTSP-based routing problem with adaptive updates, yet the method primarily focuses on tour optimization rather than explicitly preserving inspection coverage under occlusion and environmental discrepancy. PredRecon \cite{Feng2023PredRecon} accelerates aerial reconstruction by predicting surface geometry to guide viewpoint selection, but its objective is reconstruction efficiency rather than inspection completeness under partial priors. 

In contrast to these approaches, our framework assumes the availability of a reference inspection map while explicitly accounting for discrepancies between the reference map and the actual inspection environment during execution. It integrates global coverage planning with inspection-oriented local adaptation to preserve coverage quality under environmental uncertainty.

At the local planning level, recent work has investigated perception-aware local planning, where trajectory generation explicitly accounts for sensing quality and visibility during motion execution. Perception-aware full-body motion primitives \cite{PerceptionAware2024FullBodyMotionPrimitives} incorporate visibility constraints into motion primitive selection to maintain reliable perception during navigation. Global yaw optimization methods \cite{GlobalYaw2024TrajectoryOptimization} adjust camera orientation along the trajectory to improve view quality and feature observability. APACE \cite{APACEAgilePerceptionAwareTrajectory} integrates perception-aware objectives into agile trajectory optimization to enhance state estimation accuracy. TRG-Planner \cite{Lee2025TRGPlanner} further couples trajectory generation with perception-driven constraints to improve mapping and navigation performance. While these methods improve perception reliability and localization stability, their objectives are primarily navigation- or mapping-oriented. In contrast, our framework adapts the local trajectory and view angle explicitly to preserve inspection surface coverage under occlusions, directly coupling perception awareness with inspection completeness.

\begin{figure*}[htbp] 
    \centering
    \includegraphics[width=0.85\textwidth]{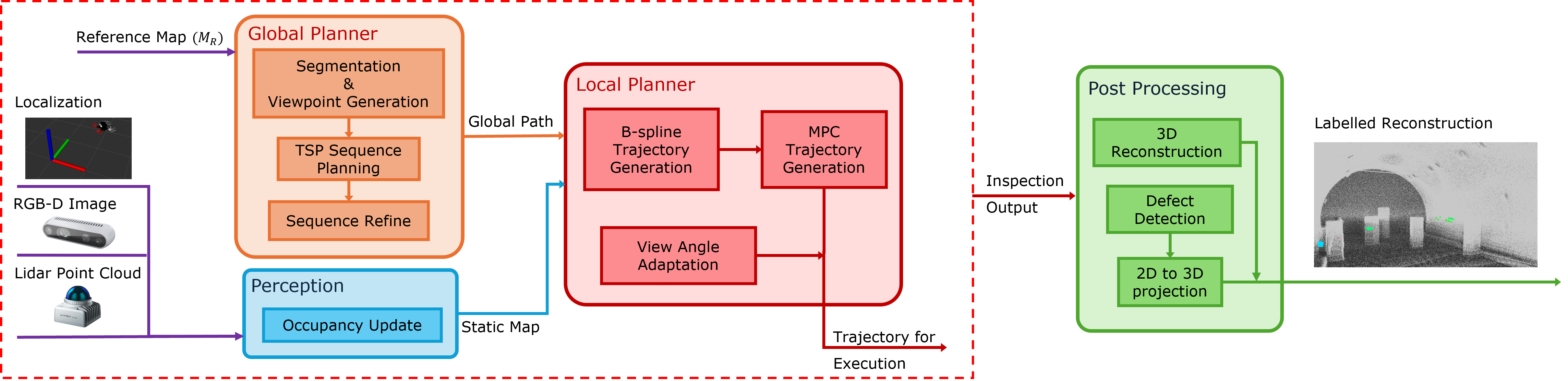}
    \caption{
    System overview of the proposed UAV surface inspection framework. 
    Given a reference map and onboard perception inputs, the global planner performs segmentation-based viewpoint generation and sequencing, while the local planner generates trajectories with online view-angle adaptation under environmental uncertainty. 
    The collected inspection data are further processed for downstream applications, including 3D reconstruction and defect detection.}
    \label{system-overview}
\end{figure*}

\section{PROBLEM FORMULATION}
This work considers surface inspection in partially known indoor environments. For each inspection task, we assume that a prior reference model, denoted as $\mathcal{M}_R$, is available. The reference map can be derived from an indoor floor plan and represents enclosed planar boundaries that are vertically extruded to a fixed height. As inspection is performed within the indoor space, only the interior-facing vertical surfaces are treated as inspection targets, while exterior-facing walls are excluded from the coverage objective.

Although $\mathcal{M}_R$ specifies the nominal geometry of the target surfaces, the real environment may contain structural deviations or additional static obstacles that are not included in the reference map. The goal is therefore to generate collision-free trajectories that maximize the coverage of the interior surfaces defined by $\mathcal{M}_R$, while adapting to environmental discrepancies during execution.

\section{METHODOLOGY}
Figure \ref{system-overview} shows an overview of the proposed framework. Providing a reference map $\mathcal{M}_R$, the goal is to efficiently achieve the maximum coverage of $\mathcal{M}_R$ using the onboard RGB-D camera for surface inspection. Localization is updated using LiDAR Inertial Odometry (LIO) 
 \cite{FASTLIO2}, and perception is provided by the onboard perception module that takes the RGB-D image and the LiDAR point cloud as inputs. The proposed planning method consists of two parts: a global planner for viewpoint generation and sequence planning (Sec. \ref{sec:global_planner}), and a local planner for collision-free trajectory generation (Sec. \ref{sec:local_planner}) and view-angle adaptation. The inspection data collected during execution can be further used for downstream tasks such as 3D reconstruction and defect detection, resulting in a labelled 3D reconstruction of the inspection environment.

\subsection{Segment-based Global Planner}\label{sec:global_planner}
% line1
The proposed global planner, as demonstrated in Alg. \ref{alg:global_planner}, takes a reference map $\mathcal{M}_R$ as input and outputs a sequence of viewpoint clusters to maximize the coverage and efficiency of surface inspection. 

% line2
The process begins with geometric segmentation of $\mathcal{M}_R$ using the Region Growing method (Line 2), which clusters the environment based on surface normals and curvature, producing a set of segments $\mathcal{S}$. 
% line3-6
For each segment $s \in \mathcal{S}$, a set of candidate viewpoints $\mathcal{V}_{\text{raw}}$ is generated to ensure comprehensive coverage (Lines 3-6). Since segmentation is based on surface normals, viewpoints within each segment are arranged in a linear spatial pattern. Each viewpoint $v_i$ is defined as a tuple $v_i = (p_i, \phi_i)$. $p_i$ denotes the position of the viewpoint, and $\phi_i$ is the view angle, which is initialized to be perpendicular to the principal orientation of the segment. Within each segment, viewpoints are spatially compact, reducing unnecessary interleaving between unrelated surfaces and simplifying the global sequencing. In addition, consecutive viewpoints within each segment naturally share similar view angles, which improves yaw consistency and increases view overlap during local execution.
% line 7
The candidate viewpoints are then sequenced using the Lin-Kernighan-Helsgaun (LKH) solver \cite{LKH} (Line 7). Then, post-processing with a three-stage refinement procedure is conducted to handle outliers and enhance consistency.
% line8-11
In the first stage of post-processing, a grouping is performed by mapping each viewpoint back to its source segment. Consecutive viewpoints from the same segment are merged into clusters, resulting in an initial cluster set $\mathcal{C}$ (Line 8). In the second stage, clusters with sizes below a threshold $\tau_{max}$ are considered outliers and merged with the nearest adjacent cluster from the same source segment, provided that their combined sequence remains contiguous (Lines 9-14). This produces a refined set of clusters $\mathcal{C}_{merged}$ that reduces fragmented visitation patterns caused by small clusters and improves the consistency of local execution. Finally, a local reordering is performed within each cluster to minimize intra-cluster travel distance while preserving the global cluster order, generating a sequence of viewpoint clusters $\mathcal{C}_{optimized}$ that serves as the basis for downstream local planning and inspection (Lines 15-16).

\begin{algorithm}[htbp]
\caption{Global Planner} 
\label{alg:global_planner}
\SetAlgoNoLine
\SetKwComment{Comment}{$\triangleright$\ }{}

$\mathcal{M}_R \gets$ \text{Input Reference Map}\;

$\mathcal{S} \gets \textbf{MapSegmentation}(\mathcal{M}_R)$ \Comment{segment bounding boxes}

$\mathcal{V}_{\text{raw}} \gets \emptyset$ \ \Comment{viewpoint array}

\For{$s \in \mathcal{S}$}{
    $\mathcal{V}_s \gets \textbf{GenerateViewpoints}(s)$ \Comment{viewpoint set for each segment}
    $\mathcal{V}_{\text{raw}}.\textbf{append}(\mathcal{V}_s)$ \;
}

$\mathcal{V}_{\text{ordered}} \gets \textbf{SolveTSP}(\mathcal{V}_{\text{raw}})$ \Comment{TSP viewpoint sequence}

$\mathcal{C} \gets \textbf{Remap}(\mathcal{V}_{\text{ordered}}, \mathcal{S})$ \Comment{cluster array}

$\tau \gets \text{Initial Threshold}$;

$\tau_{max} \gets \text{Predefined Maximum Threshold}$;

\While{$\tau<=\tau_{max}$}{
    $\mathcal{C}_{merged} \gets \textbf{MergeOutlier}(\mathcal{C}, \tau)$ \Comment{merged cluster array}
    
    $\mathcal{C} \gets \mathcal{C}_{merged}$

    $\tau \gets \tau + 1$
}
$\mathcal{C}_{optimized} \gets \textbf{LocalReorder}(\mathcal{C}_{merged}, \mathcal{S})$ \Comment{output viewpoint sequence}

\textbf{return} $\mathcal{C}_{optimized}$\;

\end{algorithm}

\subsection{Inspection-oriented Local Planner}\label{sec:local_planner}
%static and dynamic local planner
While the global planner aims to maximize the coverage and inspection efficiency of the inspection target given the reference map, the local planner is responsible for executing the global plan and adapting the motion to previously unknown obstacles encountered during execution.
The local planner integrates a static reference trajectory planner, a dynamically updated trajectory tracking and obstacle avoidance planner, and a local view angle planner to enhance navigation safety and locally improve coverage under occlusion.

% bspline static planner
\textbf{Static Planner: }
The static planner generates a reference trajectory for each cluster.
Given a global trajectory or a waypoint, a B-spline trajectory of order $k$ is constructed from a set of control points:
\begin{equation}
    \hat{\mathbb{P}} = \{P_1, P_2, \ldots, P_N\}, \quad P_i \in \mathbb{R}^3,
\end{equation}
where the first and last $k - 1$ control points are fixed to the start and goal positions, respectively. Within each cluster, the sequenced viewpoints serve as the global trajectory input. Between clusters, the first viewpoint in the next cluster serves as the waypoint input. The trajectory generation is formulated as an unconstrained optimization problem where the optimization variables are the set $\mathbb{P}$ of the intermediate $N - 2(k - 1)$ control points, excluding the fixed start and goal. The objective is to minimize a cost function composed of weighted terms for control effort, trajectory smoothness, and static collision costs:
\begin{equation}
    J(S) = \alpha_{\text{control}} \cdot J_{\text{control}} 
    + \alpha_{\text{smooth}} \cdot J_{\text{smooth}} 
    + \alpha_{\text{collision}} \cdot J_{\text{collision}}.
    \label{eq:cost_function}
\end{equation}

Details of the static planner can be found in \cite{xu2023ViGO}. The static planner serves as a reference for the downstream dynamic planner, providing a trajectory that tracks all the reachable viewpoints.

% dynamic planner
\textbf{Dynamic Planner: }
Given a reference trajectory, we apply the model predictive control in \cite{Xu2025IntentPredictionMPC} to generate final execution trajectories. The entire optimization problem is formulated as:
\begin{mini!}[2]
    {\mathbf{x}_{0:N}, \mathbf{u}_{0:N-1}}{\sum_{k=0}^{N} {\begin{Vmatrix} \mathbf{x}_{k} - \mathbf{x}_{k}^\text{ref} \end{Vmatrix}^2 + \sum_{k=0}^{N-1} \lambda_{\mathbf{u}}\begin{Vmatrix} \mathbf{u}_{k} \end{Vmatrix}^2},}{}{} \label{mpc objective}
\addConstraint{\mathbf{x}_{0}}{=\mathbf{x}(t_{0})}{} \label{initial constraint}
\addConstraint{\mathbf{x}_{k}}{=f(\mathbf{x}_{k-1}, \mathbf{u}_{k-1})}{} \label{dynamics model}
\addConstraint{\mathbf{u}_{\text{min}} \leq}{\mathbf{u}_{k} \leq  \mathbf{u}_{\text{max}}}{} \label{control limits}
\addConstraint{\mathbf{x}_{k} \not\in \mathcal{R}_{i}}{, \forall i \in \mathbb{O}}{} \label{collision constraint}
\addConstraint{\forall k \in \{0, \ldots , N\}}, 
\end{mini!}
where $\mathbf{x_k} = [\mathbf{p_k},\mathbf{v_k}]^T$ and $\mathbf{u_k} = \mathbf{a_k}$ represent the robot states and control inputs with the subscript indicating the time step. The objective (Eqn. \ref{mpc objective}) is to minimize the deviation from the reference trajectory while using the least control effort.  Equation \ref{initial constraint} sets the initial state constraint based on the current robot states. The robot's dynamics model and control limits are presented by Eqns. \ref{dynamics model} and \ref{control limits}, respectively.
The collision constraints (Eqn. \ref{collision constraint}) ensure that the robot avoids collisions with obstacles. 

\begin{figure}[t]
\centering
\includegraphics[width=\linewidth]{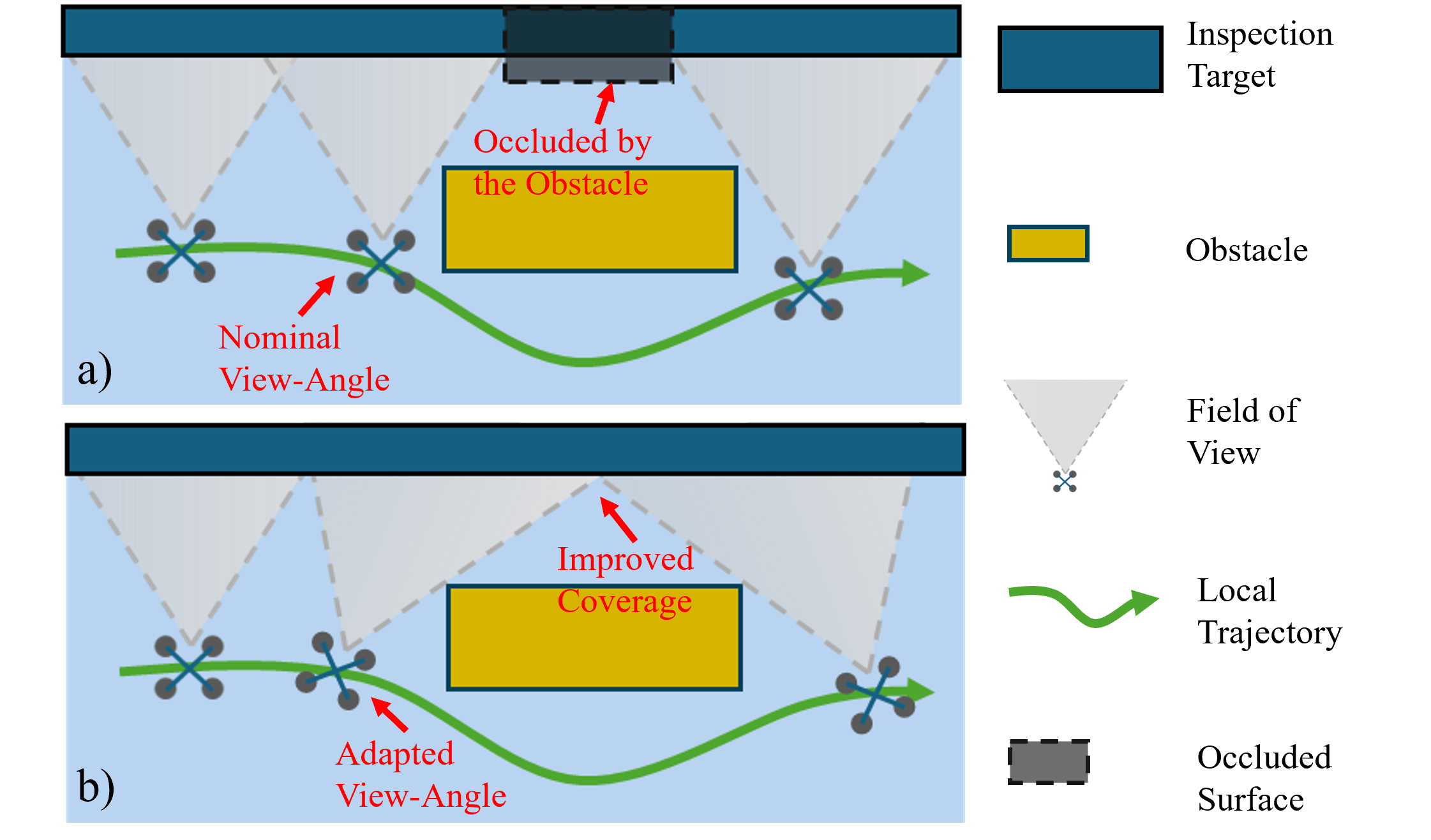}
\caption{
Illustration of the proposed view-angle adaptation strategy. 
(a) When the nominal viewing direction is strictly maintained, unforeseen obstacles may occlude portions of the inspection target, resulting in incomplete coverage. 
(b) The adaptive view-angle planner adjusts the viewing direction online to recover occluded regions while preserving the overall inspection trajectory.
}
\label{fig:view_angle_demo}
\end{figure}

\textbf{View-Angle Planner: }During execution, unforeseen obstacles may partially occlude previously planned viewpoints, leading to incomplete coverage if the nominal viewing direction is strictly maintained, as illustrated in Fig.~\ref{fig:view_angle_demo}(a). However, instead of strictly maintaining the nominal viewing direction, the system can adaptively adjust the view angle to recover occluded regions and increase coverage, as shown in Fig.~\ref{fig:view_angle_demo}(b). Therefore, we introduce an adaptive view-angle planner that adjusts the viewing direction online to maximize observable information under environmental uncertainty as demonstrated in Alg. \ref{alg:local_view_angle}.

With the reference map $\mathcal{M_R}$, a score is assigned to each grid cell in $\mathcal{M_R}$ (Lines 1-2), and these scores will be updated during the inspection process. 
During the inspection, The view-angle planner leverages the real-time occupancy map $\mathcal{M}$, incrementally updated from LiDAR point cloud data and current robot position$p_r$ to plan for optimized view angle (Lines 4-5). Given the camera field of view (FoV) and the nominal view angle of each obstructed viewpoint $\phi_i$, the corresponding occluded regions on the reference map are inferred (Lines 6-9). In addition, Grids that are observed in the real-time occupancy map $\mathcal{M}$ and belong to the reference map $\mathcal{M_R}$ are marked as scanned (Line 10). Then the scores are updated: lower scores are attributed to previously scanned regions, and higher scores are assigned to occluded or unobserved areas to reflect their observation status (Lines 11-15). Then, by utilizing the current robot state and the camera field of view, the view-angle planner performs a discrete 360-degree raycasting on the map $\mathcal{M}$ to evaluate the weighted accumulated score across candidate view angles (Lines 17-20). While the grids that are in $\mathcal{M_R}$ have the assigned score, the grids that are not in $\mathcal{M_R}$ have a score of 0. Equation \ref{equ:weight} shows the weight calculation, which makes the nominal view angle that is perpendicular to the inspection target preferable:
\begin{equation}
\label{equ:weight}
    \mathbf{w} =  (\cos{(\phi - \phi_i)} + 1)/2.
\end{equation}
The direction yielding the highest cumulative score is selected as the optimal view angle (Line 21). This adaptive mechanism enables the system to recover coverage lost to occlusions and enhances the overall robustness of the inspection process.

% TODO: review the code and revise
\begin{algorithm}[t]
\caption{Local View Angle Adaptation}
\label{alg:local_view_angle}
\SetAlgoNoLine
\SetKwComment{Comment}{$\triangleright$\ }{}

$\mathcal{M}_R \gets$ Input reference map\;
$\mathcal{W} \gets \textbf{InitializeGridScores}(\mathcal{M}_R)$ \Comment{Initialize equal scores for all grids}

\While{Inspection}{
    $\mathcal{M} \gets$ Current occupancy map\;
    $p_r \gets$ Current robot position\;

    $\mathcal{V}_{\text{blocked}} \gets \textbf{IdentifyBlockedViewpoints}(\mathcal{M})$ \Comment{Viewpoints affected by occlusions}

    \For{$v_i \in \mathcal{V}_{\text{blocked}}$}{
        $\phi_i \gets$ nominal view angle of $v_i$\;
        $\mathcal{G}_{\text{occluded}} \gets \textbf{GetOccludedRegions}(v_i, \text{FoV}, \mathcal{M}_R)$\;
    }
    $\mathcal{G}_{\text{scanned}} \gets \mathcal{M} \cap \mathcal{M_R}$\;

    \For{$g \in \mathcal{M}_R$}{
        \If{$g \in \mathcal{G}_{\text{occluded}}$}{
            $\mathcal{W}[g] \gets \textbf{IncreaseScore}()$\;
        }
        \ElseIf{$g \in \mathcal{G}_{\text{scanned}}$}{
            $\mathcal{W}[g] \gets \textbf{DecreaseScore}()$\;
        }
    }

    $\Phi \gets$ Candidate view angles\;

    \For{$\phi \in \Phi$}{
        $S[\phi] \gets \textbf{RaycastScore}(\phi, p_r, \text{FoV}, \mathcal{W}, \mathcal{M}, \mathcal{M_R})$\ \Comment{Sum of visible grid scores}
        $\mathbf{w} \gets \textbf{GetWeight}(\phi_i)$\;
        $S_w[\phi] \gets \mathbf{w}^\top S[\phi]$\ \Comment{Weighted sum of visible grid scores}
    }

    $\phi^* \gets \arg\max_{\phi \in \Phi} S_w[\phi]$\;
    \textbf{return} $\phi^*$\;
}
\end{algorithm}

\subsection{Post Process} \label{sec:post_processing}
Beyond trajectory planning, the collected data can be further utilized for downstream infrastructure assessment tasks such as 3D reconstruction and defect localization.

For 3D reconstruction, LiDAR measurements are accumulated using a LiDAR-inertial odometry pipeline \cite{FASTLIO2}, producing a dense geometric representation of the inspected environment.

To extract surface-level defect information, we perform image-based post-processing on the inspection data. Specifically, Segment Anything (SAM) \cite{SAM} is used to obtain high-level image segmentation, providing structured region proposals of the inspected surfaces. Within each segmented region, YOLO-based detection \cite{YOLO} is applied to identify potential surface defects such as cracks. The detected defects are then projected into the reconstructed 3D space using calibrated sensor extrinsics, producing a 3D reconstruction with crack annotations.

This module is independent of the core planning algorithm and serves to demonstrate that the inspection framework generates data suitable for practical downstream analysis tasks.

\section{RESULTS AND DISCUSSIONS}
To evaluate the proposed method, we conduct simulations
and flight tests in different indoor environments. An Intel RealSense D435i camera and a Livox Mid-360 LiDAR are used for the perception. Physical flight tests run on an NVIDIA Orin
NX onboard computer. 

\subsection{Simulation Experiments}
\begin{figure}[t]
  \centering
  \includegraphics[width=\linewidth]{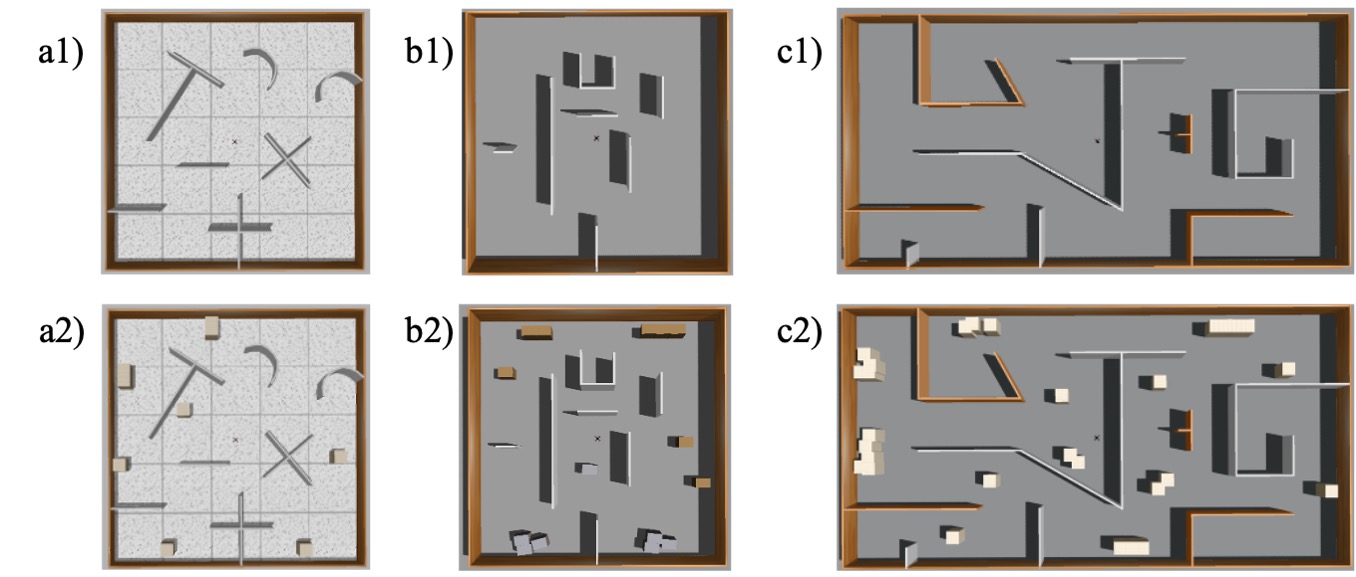}
  \caption{
    Simulated inspection environments used for qualitative evaluation. 
    (a1, b1, c1): Three different reference indoor surface layouts. 
    (a2, b2, c2): Corresponding environments with additional randomly placed obstacles to simulate partial environmental knowledge.
    }
  \label{fig:sim_env}
\end{figure}

We first qualitatively evaluate the overall behavior of the proposed framework in simulated indoor environments. 
Three representative layouts with different geometric configurations are considered, as shown in Fig.~\ref{fig:sim_env}. (a1), (b1), and (c1) show the inspection targets, serving as the reference map for the inspection framework. 
To simulate partial environmental knowledge, additional static obstacles are randomly placed within the reference layouts.

\begin{figure}[htbp]
  \centering
  \includegraphics[width=\linewidth]{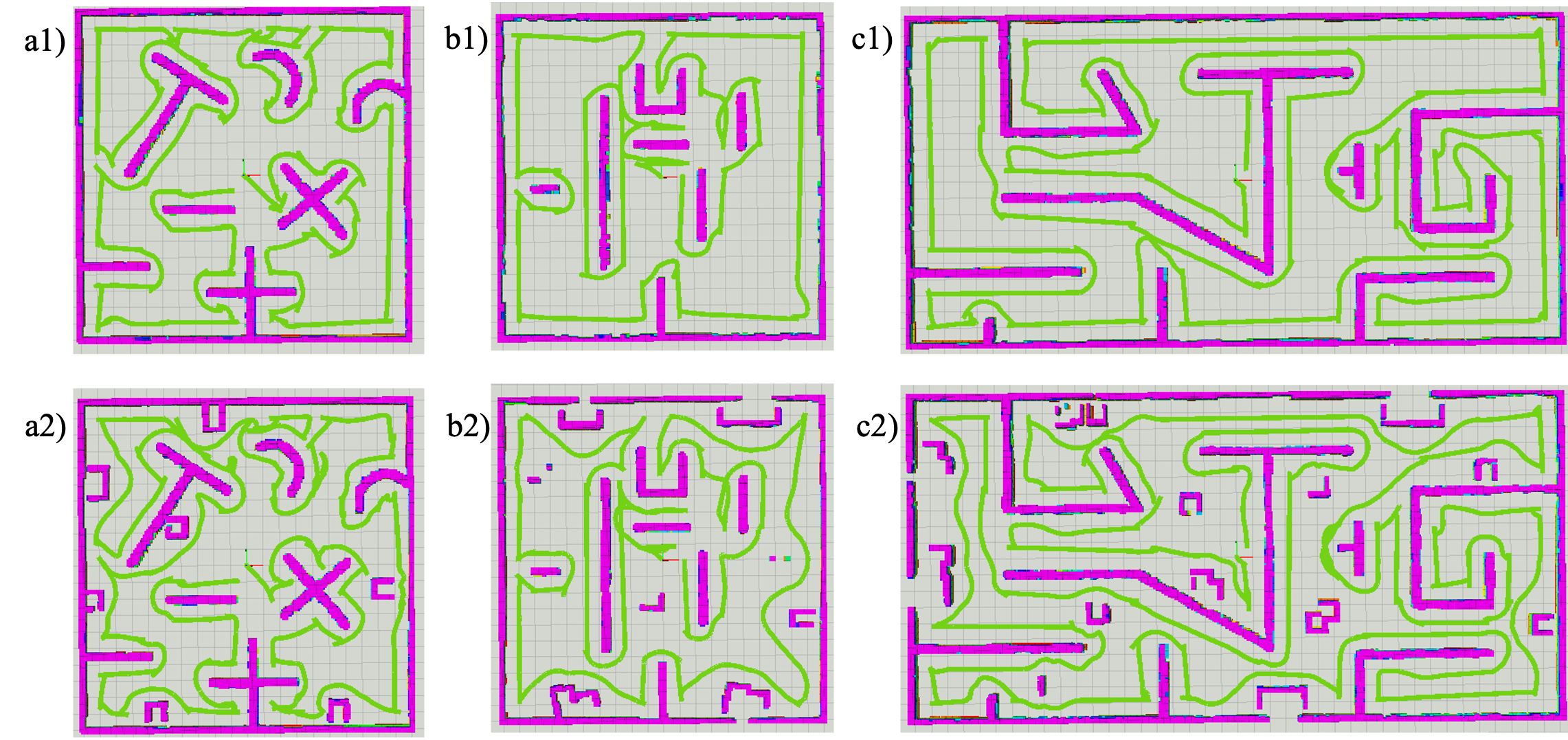}
  \caption{Qualitative results of the entire framework in three representative simulated environments. 
(a1, b1, c1): The coverage and trajectory without additional obstacles. 
(a2, b2, c2): The results under randomly placed obstacles. 
Green curves denote the inspection trajectories, and colored grids indicate inspected surface regions.}
  \label{fig:sim_result}
\end{figure}

Figure~\ref{fig:sim_result} presents the corresponding inspection trajectories and surface coverage results. 
In obstacle-free scenarios, which are shown in (a1), (b1) and (c1), the segment-based global planner generates structured and compact coverage patterns aligned with surface geometry. 
When obstacles are introduced, which are shown in (a2), (b2) and (c2), the local planner adapts the execution trajectory and viewing direction to accommodate environmental changes while preserving the overall coverage structure.

These qualitative results illustrate that the proposed framework maintains systematic surface coverage under both nominal and perturbed conditions, forming the basis for the quantitative evaluations presented in the following sections.

\subsection{Global Coverage Planner Evaluation}
\begin{table*}[t]
\centering
\caption{Benchmark comparison of global planning performance across five scene configurations (mean $\pm$ std over 10 randomized instances per configuration). Cov.: coverage (\%), Len.: trajectory length (m), $|\Delta \mathrm{Yaw}|$: mean absolute yaw change (rad).}
\label{tab:global_planner_results}
\setlength{\tabcolsep}{6pt} % adjust column padding if needed (e.g., 4--8pt)
\renewcommand{\arraystretch}{1.15} % row height
\begin{tabular}{llccccc}
\toprule
Method & Metric & Config1 & Config2 & Config3 & Config4 & Config5 \\
\midrule

\multirow{3}{*}{Ours}
& Cov. (\%) & \textbf{99.86}$\pm$0.09 & 99.84$\pm$0.17 & \textbf{99.61}$\pm$0.37 & 99.64$\pm$0.34 & \textbf{99.63}$\pm$0.23 \\
& Len. (m)  & \textbf{218.85}$\pm$27.67 & \textbf{208.71}$\pm$16.21 & \textbf{262.20}$\pm$19.50 & \textbf{252.52}$\pm$21.39 & \textbf{290.81}$\pm$30.79 \\
& $|\Delta \mathrm{Yaw}|$ (rad)
           & \textbf{0.39}$\pm$0.05 & \textbf{0.49}$\pm$0.05 & \textbf{0.60}$\pm$0.04 & \textbf{0.48}$\pm$0.03 & \textbf{0.39}$\pm$0.05 \\
\midrule

\multirow{3}{*}{FC-Planner~\cite{Feng2024FCPlanner}}
& Cov. (\%) & 99.82$\pm$0.32 & \textbf{99.92}$\pm$0.10 & 99.60$\pm$0.56 & 99.65$\pm$0.72 & 99.49$\pm$0.45 \\
& Len. (m)  & 359.73$\pm$37.15 & 312.06$\pm$33.05 & 424.67$\pm$75.75 & 381.99$\pm$47.79 & 395.91$\pm$56.80 \\
& $|\Delta \mathrm{Yaw}|$ (rad)
           & 1.38$\pm$0.10 & 1.35$\pm$0.08 & 1.37$\pm$0.06 & 1.34$\pm$0.05 & 1.38$\pm$0.08 \\
\midrule

\multirow{3}{*}{HCPP~\cite{HCPP}}
& Cov. (\%) & \textbf{99.86}$\pm$0.20 & 99.84$\pm$0.19 & 99.53$\pm$0.41 & \textbf{99.71}$\pm$0.27 & 99.54$\pm$0.31 \\
& Len. (m)  & 353.37$\pm$39.51 & 342.62$\pm$32.37 & 423.26$\pm$39.01 & 379.02$\pm$36.22 & 417.01$\pm$38.34 \\
& $|\Delta \mathrm{Yaw}|$ (rad)
           & 1.32$\pm$0.08 & 1.38$\pm$0.12 & 1.40$\pm$0.05 & 1.35$\pm$0.08 & 1.37$\pm$0.08 \\
\bottomrule
\end{tabular}
\end{table*}

We quantitatively evaluate the proposed global planner across five scene configurations. 
Each configuration is generated within a fixed boundary size and composed of different numbers and combinations of structural components. 
The boundary creates an enclosed structure to simulate indoor environments. The structural components include line, L-shaped, T-shaped, cross-shaped, arc, and circular segments, representing common geometric primitives encountered in structured indoor environments such as corridors, intersections, and curved boundaries.
For each scene configuration, the number of components ranges from 10 to 15, and 10 randomized instances are created by varying the spatial placement of components inside the boundary. 
For each instance, coverage, trajectory length, and mean absolute yaw change between consecutive viewpoints $(|\Delta Yaw|)$ are computed to evaluate coverage quality, path efficiency, and orientation consistency, respectively.
The reported results correspond to the mean and standard deviation over the 10 randomized instances per configuration.

We compare the proposed global planner against two representative baselines: FC-Planner~\cite{Feng2024FCPlanner} and HCPP~\cite{HCPP}. 
FC-Planner formulates coverage planning by decomposing the environment into subspaces and generating viewpoints via skeleton guidance, followed by global path optimization to connect viewpoints efficiently. 
HCPP is a hierarchical coverage planning framework that computes viewpoint sequences and coverage paths based on the connectivity structure of the environment. 
Both baselines are designed for deterministic coverage planning given a prior map and have been applied in general inspection tasks, making them suitable references for benchmarking global sequencing performance.

Table~\ref{tab:global_planner_results} summarizes the quantitative results across five configurations. 
All methods achieve nearly complete coverage, with differences remaining marginal across configurations. 
This indicates that the proposed segmentation and sequencing strategy does not compromise coverage quality.
In terms of trajectory efficiency, the proposed method consistently produces shorter paths compared with the two baseline methods across all scenes. 
The reduction in trajectory length suggests that organizing viewpoints into coherent surface-aligned clusters yields more compact global sequencing.
In addition, the proposed global planner generates substantially lower mean absolute yaw change $(|\Delta Yaw|)$ across all configurations. 
This demonstrates improved orientation consistency between consecutive viewpoints, reflecting smoother execution behavior and more view overlap while maintaining coverage performance.
Overall, the results indicate that the proposed segment-based global planner improves execution-level efficiency and orientation consistency without sacrificing coverage quality.

\subsection{Local View-Angle Adaptation Evaluation}
\begin{figure}[t]
\centering
\includegraphics[width=\linewidth]{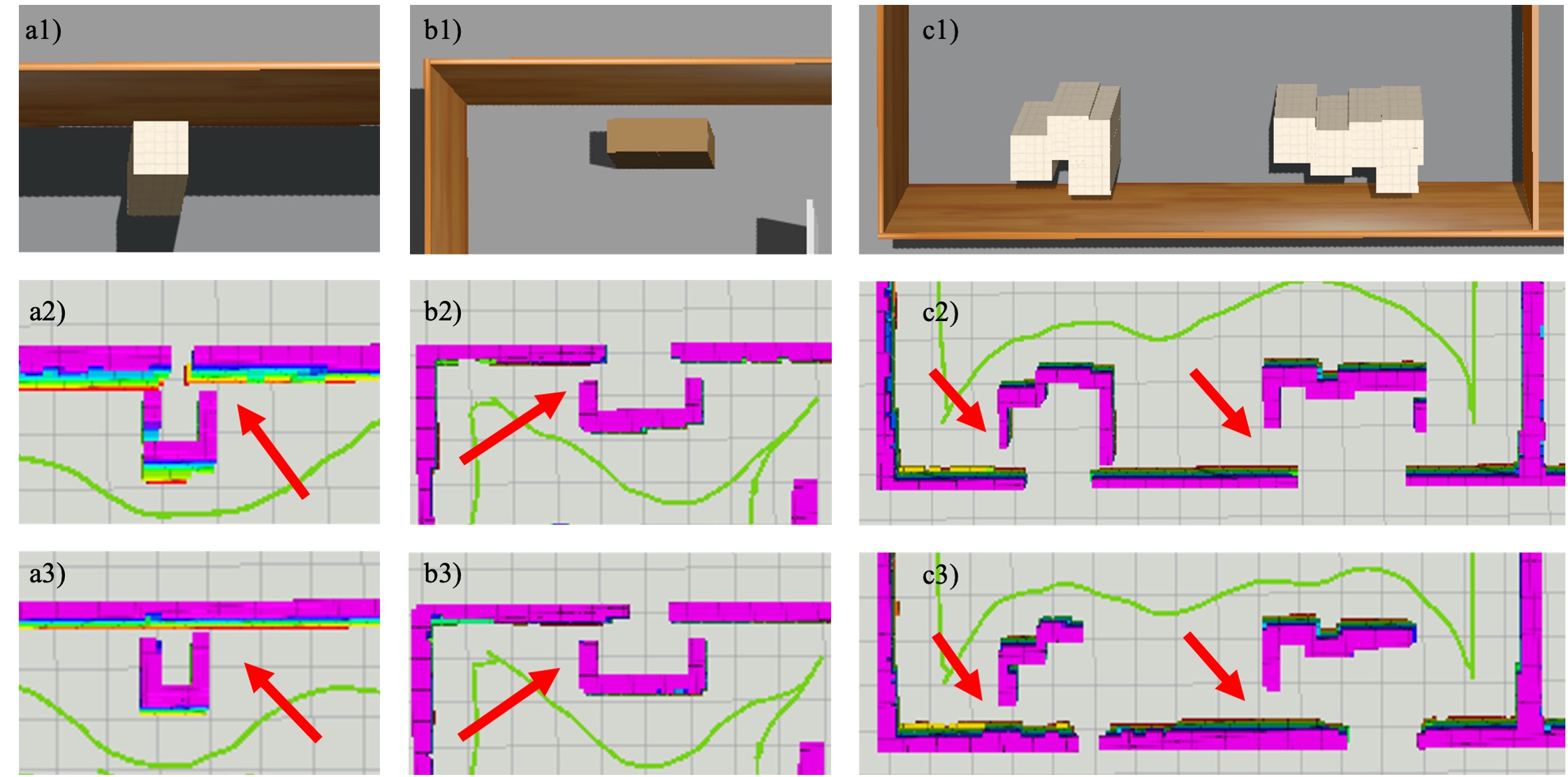}
\caption{
Qualitative comparison of local view-angle adaptation under occlusion. Occluded regions are marked using red arrows.
(a1, b1, c1): Simulation environments with increasing structural complexity and obstacle configurations. 
(a2, b2, c2): Inspection results without view-angle adaptation, where the nominal view-angle generated by the global planner is maintained. 
(a3, b3, c3): Inspection results with the proposed adaptive view-angle planner. 
Adaptive view-angle adjustment increases observable surface regions under occlusions.
}
\label{fig:local_result}
\end{figure}

We first qualitatively evaluate the effect of view-angle adaptation under occlusion. 
Figure~\ref{fig:local_result} presents three representative scenarios where the inspection target is occluded by obstacles.
When the nominal viewing direction is strictly maintained (Fig.~\ref{fig:local_result} (a2, b2, c2)), occlusions lead to partially unobserved surface regions. 
In contrast, enabling the proposed adaptive view-angle planner (Fig.~\ref{fig:local_result} (a3, b3, c3)) increases the observable surface coverage by adjusting the viewing direction online while preserving the planned trajectory structure.

To quantitatively evaluate the effect of view-angle adaptation, we focus on the coverage of the inspection target in occluded scenarios. 
Cases where obstacles completely overlap with inspection targets or do not occlude any inspection region are excluded from evaluation, as they do not reflect view-angle adjustment scenarios.
We generate over 20 occlusion scenarios with varying obstacle placements. 
For each scenario, occluded regions are defined by projecting obstacles onto the inspection target under the nominal viewing direction. 
Coverage is then computed specifically within these occluded regions.
We compare the coverage ratio within occluded regions under two settings: 
(i) maintaining the nominal view-angle, and 
(ii) enabling adaptive view-angle adjustment. 
The reported results correspond to the mean and standard deviation across all occlusion scenarios.
Within the occluded regions, the average coverage ratio without view-angle adaptation is $0.27 \pm 0.16$, while enabling the adaptive view-angle planner increases the coverage ratio to $0.42 \pm 0.24$. 
This improvement demonstrates that online view-angle adjustment effectively increases observable surface regions under occlusion.

% TODO: add Mill19 2nd floor result
\subsection{Physical Flight Test}
\begin{figure}[t]
  \centering  \includegraphics[width=\linewidth]{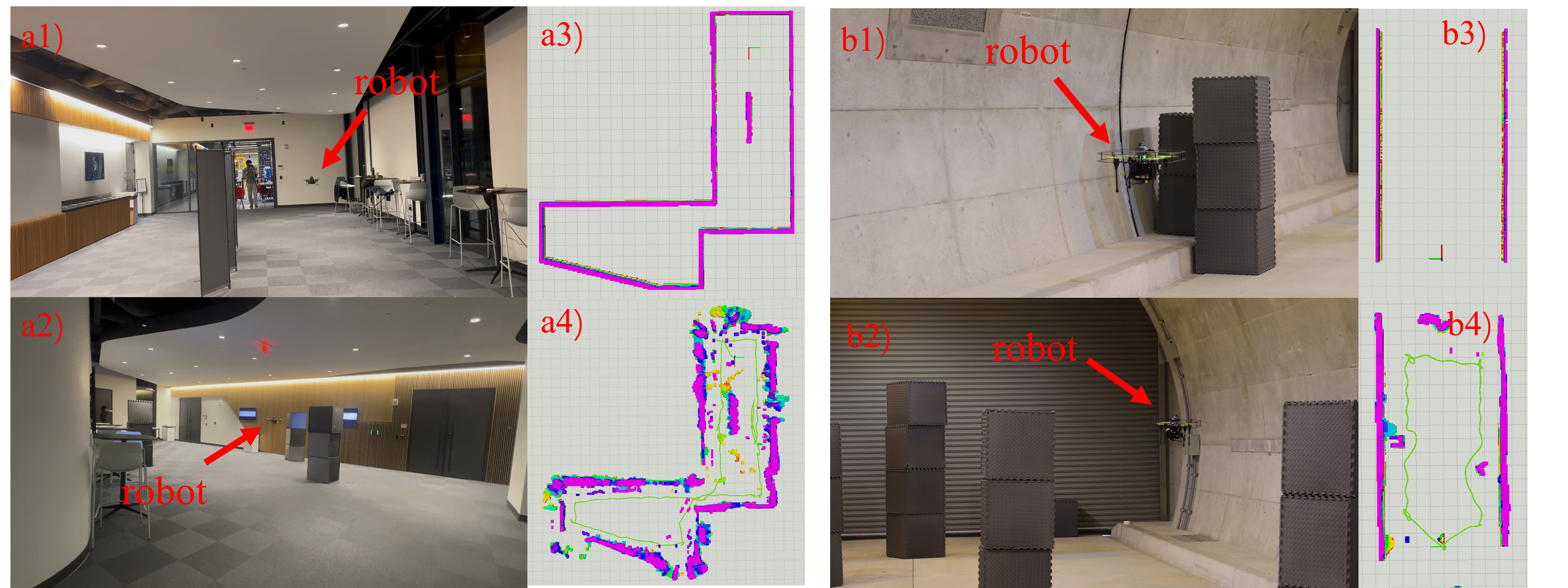}
  \caption{Physical flight test in an indoor environment. (a1 - a4) are showing our UAV inspecting in an office building. (b1-b4) are showing our UAV inspecting in a tunnel.
(a1,a2,b1,b2): Overview of the inspection environment and the UAV performing autonomous inspection in the environment.
(a3, b3): Reference map of the inspection target. 
(a4, b4): Inspection trajectory and surface coverage obtained during flight with onboard RGB-D camera.}
  \label{fig:result_combined}
\end{figure}

\begin{figure}[t]
  \centering  \includegraphics[width=\linewidth]{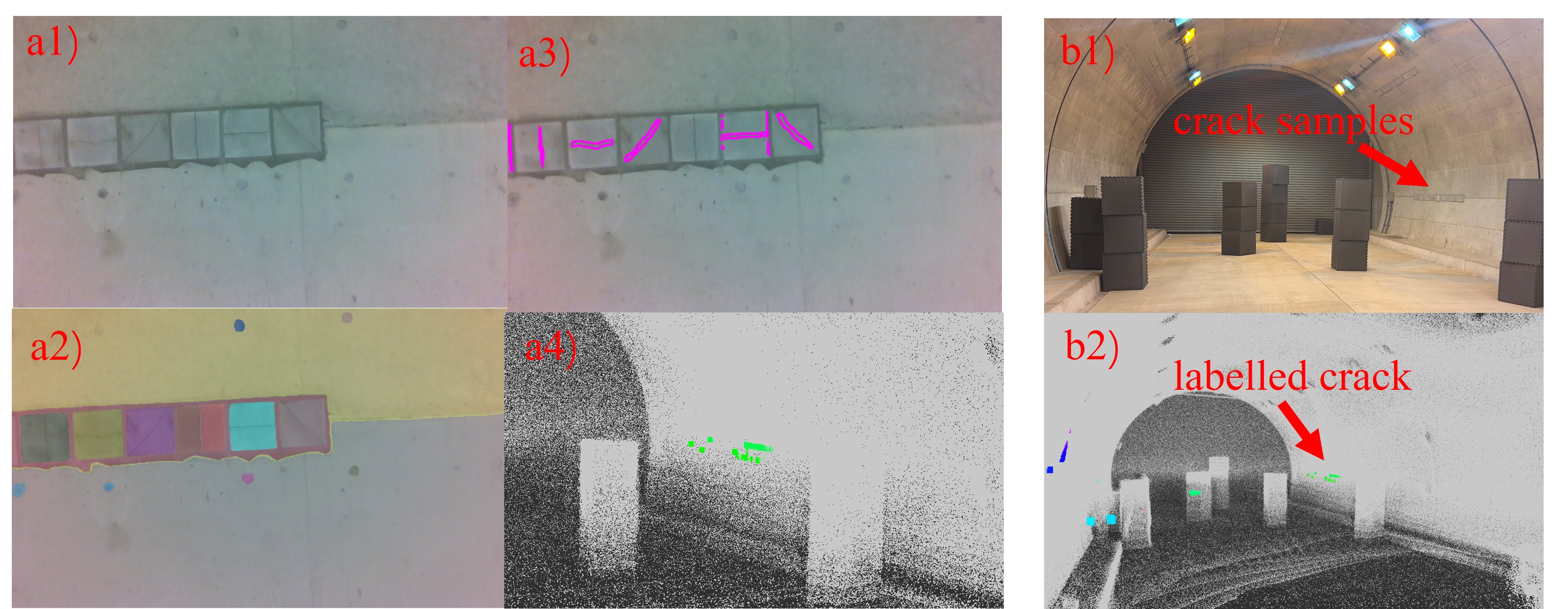}
  \caption{Post-processing results using inspection data collected during the tunnel flight experiment. (a1-a4) are showing the crack detection and projection procedure. (b1-b2) are showing the 3D reconstruction result.
(a1) Raw inspection image from onboard camera. 
(a2) Surface segmentation results. 
(a3) Detected cracks. 
(a4) Projection of cracks to 3D. 
(b1) Tunnel inspection environment with crack samples. 
(b2) 3D reconstruction of the tunnel with the labeled cracks. 
These results demonstrate that the collected inspection data can support downstream tasks such as reconstruction and defect localization.}
  \label{fig:post_process}
\end{figure}
% \begin{figure}[t]
%   \centering  \includegraphics[width=\linewidth]{images/tunnel_flight.png}
%   \caption{Physical flight test in a tunnel with post-processing demonstration. 
% (a) Tunnel inspection environment with unforeseen obstacles. 
% (b) Reference map of the inspection target. 
% (c) Inspection trajectory and accumulated surface coverage during flight using an RGB-D camera. 
% (d) UAV performing autonomous inspection. 
% (e) Example defect detection results on collected inspection images during the flight using the onboard camera. 
% (f) Reconstructed point cloud obtained from onboard LiDAR with position of potential defects labelled.}
%   \label{fig:result_tunnel}
% \end{figure}
We conduct real-world flight experiments to validate the proposed inspection framework in structured indoor environments. The complete inspection process for the real-world experiments is provided in the supplementary video.

Figure~\ref{fig:result_combined} illustrates the deployment of the system in an office building with planar wall surfaces and moderate structural variations. 
(a1) and (a2) show our UAV conducting autonomous inspection in the office building environment, which contains unknown obstacles compared to the reference map of the inspection target shown in (a3).
During flight, the occupancy map is incrementally updated using onboard LiDAR and RGB-D measurements, and the UAV follows the globally planned viewpoint sequence while executing local adaptive trajectory planning.
(a4) shows the coverage from the onboard camera view, while the executed trajectory remains consistent with the planned inspection layout.  
Missing coverage regions are mainly caused by windows and transparent doors, which cannot be reliably observed by the RGB-D sensor.
The experiment demonstrates that the proposed planning framework can be executed in real indoor environments while preserving inspection continuity.

The second flight experiment is conducted in a tunnel to demonstrate the applicability of the proposed framework in infrastructure inspection tasks.
The tunnel environment contains structural obstacles that are not included in the reference map, introducing partial occlusions during inspection.
As illustrated in Fig.~\ref{fig:result_combined} (b1-b4), the UAV autonomously inspects the tunnel wall surface while navigating around unforeseen obstacles. 
The accumulated surface coverage shown in Fig.~\ref{fig:result_combined} (b4) indicates that inspection observations are continuously collected along the tunnel walls.

To further demonstrate the usability of the collected inspection data, images captured during autonomous flight are processed for defect detection following the post-processing procedure described in Sec.~\ref{sec:post_processing}. Figure~\ref{fig:post_process} presents representative results from the tunnel inspection experiment. 
Figures~\ref{fig:post_process} (a1–a4) illustrate crack detection on an example RGB image collected during inspection. The raw image (a1) is first segmented using SAM to obtain structured surface regions (a2), enabling more localized defect analysis. Crack regions detected by the YOLO-based model (a3) are subsequently projected into the reconstructed 3D space (a4) using camera extrinsics. 
In addition, the LiDAR-based reconstructed point cloud is shown in Fig.~\ref{fig:post_process} (b2), where detected defect locations are visualized using colored grids. The annotated defect positions correspond to the crack samples placed in the tunnel environment. 
These results indicate that the inspection data generated by the proposed framework are suitable for downstream reconstruction and defect analysis tasks.

\section{CONCLUSION AND FUTURE WORK}
This work presents an adaptive UAV inspection framework that integrates segment-based global coverage planning with inspection-oriented local view-angle adaptation. 
The global planner generates an efficient and complete inspection sequence based on a reference map, while the local planner adjusts the viewing direction online to mitigate coverage loss due to occlusions during execution. 
Extensive simulation results demonstrate that the proposed method maintains near-complete coverage while reducing trajectory length and orientation variation. 
Real-world flight experiments further validate that the framework can be executed in structured indoor environments and produce usable inspection data for downstream analysis.
The current framework focuses on structured indoor environments, where inspection targets can be represented by floorplan-derived, vertically extruded planar surfaces. Extending the approach to arbitrary free-form 3D surfaces would require more general surface decomposition and viewpoint sampling strategies. Future work will focus on more complex structural geometries and tighter coupling between global planning and online perception for improved adaptability.
\addtolength{\textheight}{-12cm}   % This command serves to balance the column lengths

%%%%%%%%%%%%%%%%%%%%%%%%%%%%%%%%%%%%%%%%%%%%%%%%%%%%%%%%%%%%%%%%%%%%%%%%%%%%%%%%

%%%%%%%%%%%%%%%%%%%%%%%%%%%%%%%%%%%%%%%%%%%%%%%%%%%%%%%%%%%%%%%%%%%%%%%%%%%%%%%%
\bibliographystyle{IEEEtran}
\bibliography{bibliography}

\end{document}